\documentclass{article} 
\usepackage{iclr2025_conference,times}

\usepackage{amsmath,amsfonts,bm}

\def\eqref#1{equation~\ref{#1}}

\def\1{\bm{1}}

\DeclareMathAlphabet{\mathsfit}{\encodingdefault}{\sfdefault}{m}{sl}
\SetMathAlphabet{\mathsfit}{bold}{\encodingdefault}{\sfdefault}{bx}{n}

\usepackage{url}
\usepackage{lipsum}
\usepackage{natbib}
\usepackage[colorlinks=true,citecolor=apolloblue,linkcolor=apolloblue,urlcolor=apolloblue]{hyperref}
\usepackage{microtype}
\usepackage{graphicx}
\usepackage{booktabs} 
\usepackage{longtable}
\usepackage{tabularx}
\usepackage{subcaption}
\usepackage{amsmath}
\usepackage{listings}
\usepackage{float}

\usepackage{amsmath}
\usepackage{amssymb}
\usepackage{mathtools}
\usepackage{amsthm}
\usepackage{algorithmic}

\graphicspath{ {./figures/} }

\title{Forecasting Frontier Language Model Agent Capabilities}

\author{
Govind Pimpale\thanks{Equal contribution}\\\textmd{MATS}
\And
Axel H{\o}jmark\footnotemark[1]\\\textmd{MATS \& Apollo Research}
\AND
J\'er\'emy Scheurer\thanks{Equal contribution}\\\textmd{Apollo Research}
\And
Marius Hobbhahn\footnotemark[2]\\\textmd{Apollo Research}
}

\iclrfinalcopy 
\begin{document}

\maketitle

\vspace{0.3in} 
\begin{abstract}
    As Language Models (LMs) increasingly operate as autonomous agents, accurately forecasting their capabilities becomes crucial for societal preparedness. 
    We evaluate six forecasting methods that predict downstream capabilities of LM agents. 
    We use ``one-step'' approaches that predict benchmark scores from input metrics like compute or model release date directly or ``two-step'' approaches that first predict an intermediate metric like the principal component of cross-benchmark performance (PC-1) and human-evaluated competitive Elo ratings.
    We evaluate our forecasting methods by backtesting them on a dataset of 38 LMs from the OpenLLM 2 leaderboard. 
    We then use the validated two-step approach (Release Date$\to$Elo$\to$Benchmark) to predict LM agent performance for frontier models on three benchmarks:
    SWE-Bench Verified (software development), Cybench (cybersecurity assessment), and RE-Bench (ML research engineering). 
    Our forecast predicts that by the beginning of 2026, non-specialized LM agents with low capability elicitation will reach a success rate of 54\% on SWE-Bench Verified, while state-of-the-art LM agents will reach an 87\% success rate.
    Our approach does not account for recent advances in inference-compute scaling and might thus be too conservative.
\end{abstract}

\begin{figure*}
    \centering
    \includegraphics[width=\textwidth]{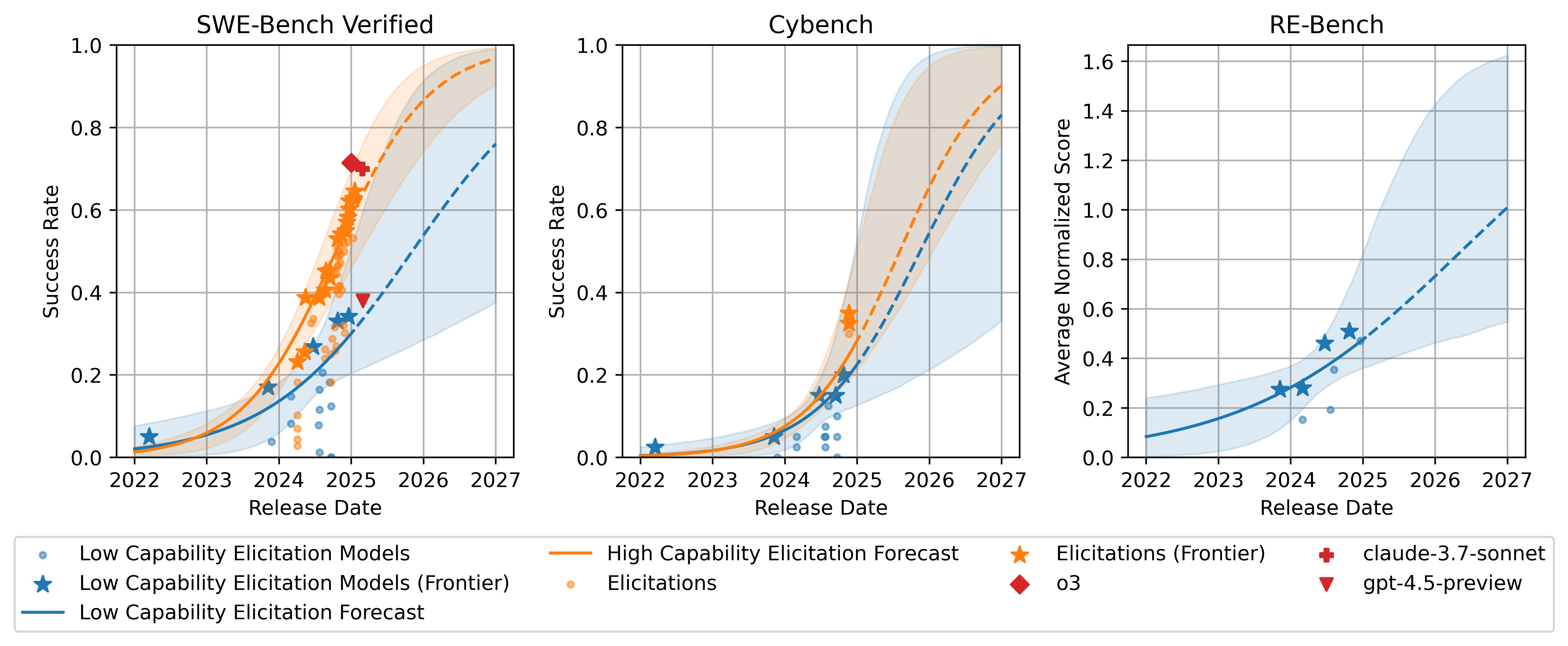}
    \vskip -0.1in
    \caption{Low-Elicitation and High-Elicitation forecasts for LM agent performance on SWE-Bench, Cybench, and RE-Bench. Elicitation level refers to performance improvements from optimizing agent scaffolds, tools, and prompts to achieve better results. Forecasts are generated by predicting Chatbot Arena Elo-scores from release date and then benchmark score from Elo. The low-elicitation (blue) forecasts serve as a conservative estimate, as the agent has not been optimized and does not leverage additional inference compute. 
    The high-elicitation (orange) forecasts use the highest publicly reported performance scores. Because RE-Bench has no public high-elicitation data, it is excluded from these forecasts.
    } 
    \label{fig:scaling-graph}
\end{figure*}

\section{Introduction}
Large language models are increasingly trained and deployed as autonomous agents capable of independently pursuing goals and executing longer-form real-world tasks. This rapid advancement creates a need to forecast when these systems will reach critical capability thresholds in order to understand and prepare for their effects on society and the economy.

While there are several approaches that forecast LM capabilities, they face key limitations. Methods like Observational Scaling Laws \citep{ruan2024observational} and Sloth \citep{polo2024sloth} can only make predictions after evaluating models from a new model family on multiple benchmarks to derive their predictive metrics, such as PC-1 scores (which correspond directly to general capabilities) \citep{ruan2024observational} or family-specific efficiency parameters (which can be used to convert FLOP to general capabilities) \citep{polo2024sloth}\footnote{While Observational Scaling Laws \citet{ruan2024observational} allows direct FLOP-based predictions for specific model families, this approach fails to account for algorithmic advances across families and would therefore underestimate future benchmark performance.}. Other approaches that predict directly from compute \citep{owen2024predictablelanguagemodelbenchmark} or release date are either less accurate \citep{ruan2024observational} or have not been systematically evaluated.

Backtesting these six methods, we find that the two-step approach with a linear relationship between date and Elo and a sigmoidal relationship from Elo to benchmark performs competitively and has publicly available data for frontier models. 
Our forecasts focus on \emph{frontier} performance, the performance of the best-known model at a given time or compute level, unlike previous work that focuses on predicting average model performance \citep{ruan2024observational, polo2024sloth}. This is relevant because frontier performance determines what capabilities can be automated at a given time and because frontier models are most likely to present novel risks and societal impacts \citep{hendrycks2023overviewcatastrophicairisks}. Finally, our method only requires model release dates and Elo Ratings, information that is typically public even for proprietary models, rather than metrics like FLOP count, parameter count, or dataset size, which are often unavailable for frontier models.

Our \textbf{contributions} are as follows:
\begin{itemize}
    \item We show that Chatbot Arena Elo \citep{chatbot_arena} is a good proxy for a model's underlying performance and correlates strongly with FLOP count.
    \item We compare 6 methodologies that predict performance based on combinations of FLOP count, model release date, Elo, and PC-1. We backtest all methods on 38 models from Open LLM leaderboard 2 \citep{open-llm-leaderboard-v2}, which have Elo scores available, and find that Release Date$\to$PC-1$\to$Benchmark performs best closely followed by Release Date$\to$Elo$\to$Benchmark.
    \item We evaluate 17 models on SWEBench and Cybench with a simple agent scaffold and use publicly available results of RE-Bench \citep{wijk2024rebenchevaluatingfrontierai} to enable direct performance comparisons across models. 
    Our low-elicitation prediction projects that non-specialized LM agents will achieve a 55\% success rate on SWE-Bench Verified tasks by early 2026, while our high-elicitation forecast suggests 85\%. \footnote{All code for the analysis, scaffold, and evaluation is available at \href{https://github.com/pimpale/forecasting-frontier-language-model-agent-capabilities}{https://github.com/pimpale/forecasting-frontier-language-model-agent-capabilities}}

\end{itemize}

\section{Methods}

\begin{figure*}
    \centering
    \includegraphics[width=0.9\textwidth]{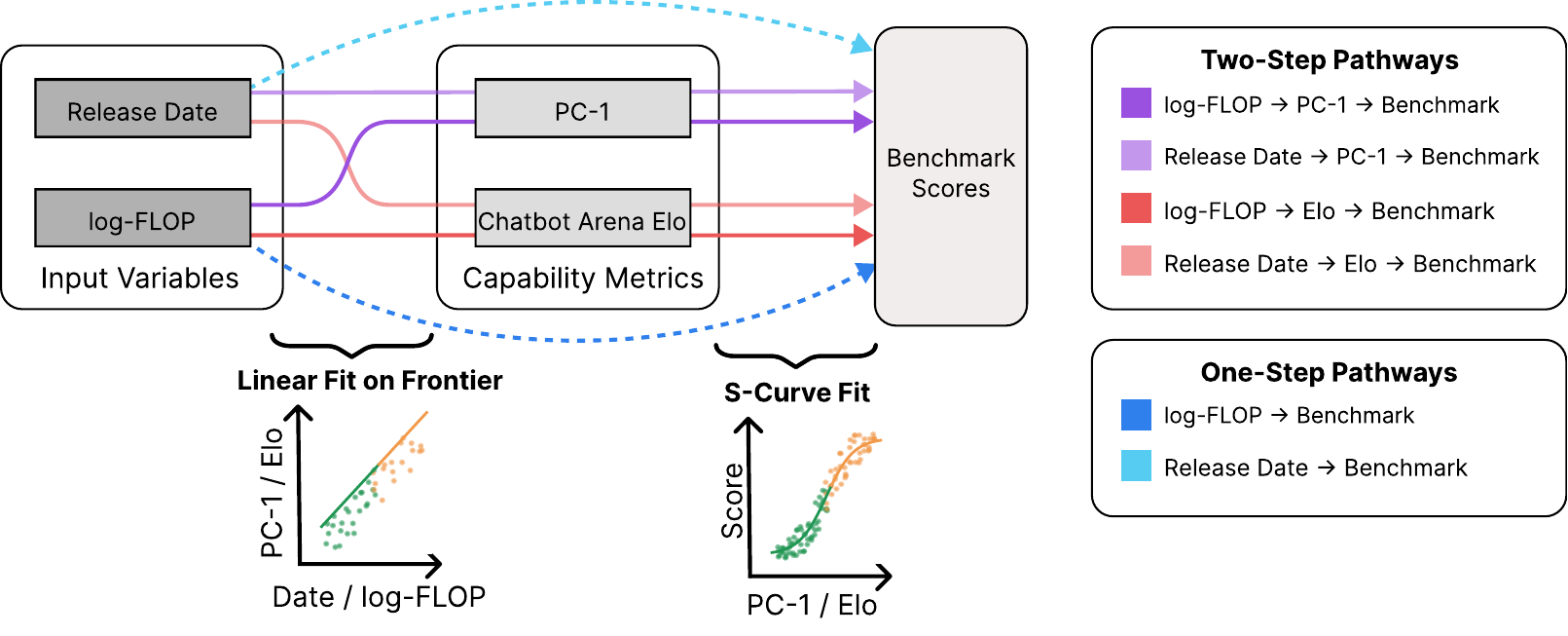}
    \vskip -0.1in
    \caption{
     Six approaches for predicting frontier LM capabilities. Two direct methods (blue pathways) model benchmark performance as a sigmoid function of either release date or compute (log-FLOP). Four two-step methods (red and purple pathways) first use a linear function to predict intermediate capability metrics (PC-1 or Chatbot Arena Elo) from input variables, then map these metrics to benchmark scores using a sigmoid function.
    }
    \label{fig:paths}
\end{figure*}

We evaluate six different approaches to predict frontier LM capabilities.
The simplest approaches try to directly predict frontier benchmark performance from an \emph{input variable} such as training log-FLOP or release date. 

We contrast such one-step approaches with two-step approaches where we first use the input variable to predict an intermediate \emph{capability metric} (e.g. PC-1 or Elo), and then use that capability metric to forecast the downstream \emph{benchmark performance}. These variables can be combined in four different ways. When combined with the one-step approaches, this results in six different methods to forecast the performance on a target benchmark (see Figure \ref{fig:paths}).

\subsection{Prediction Tasks}
In the following, we describe which input variables and downstream benchmarks we use.

\subsubsection{Input Variables}

Input variables are broad and general quantities that we expect to have predictive power for downstream performance. They don't require knowing any specifics about the model, e.g. architecture or exact training procedure. 

\textbf{Scaled training log-FLOP}: The amount of compute used to train the model measured in FLOP. Previous literature has found an approximately log-linear relationship between input compute and model performance \citep[e.g.][]{Finnveden_2020, owen2024predictablelanguagemodelbenchmark}. \citet{sevilla2022computetrends} observes that the amount of compute utilized by large scale pre-training runs tends to double approximately once every 9 months, providing us with a broad reference class for the compute requirements of frontier model training.

In this paper, we always use \emph{scaled} FLOP as described in \citet{owen2024predictablelanguagemodelbenchmark}. It is a common practice to ``overtrain'' models \citep{shafkat2023, dubey2024llama3herdmodels} by reducing parameter count and increasing dataset size beyond what would be optimal under Hoffman scaling laws \citep{hoffmann2022trainingcomputeoptimallargelanguage} to reduce cost at inference time. Therefore, raw (unscaled) FLOP estimates are less comparable. We can overcome this by normalizing all models to the lowest possible FLOP count that would achieve the same loss using Hoffman scaling laws. See Appendix~\ref{app:scaled_compute} for details.

\begin{figure*}
    \centering
    \includegraphics[width=\textwidth]{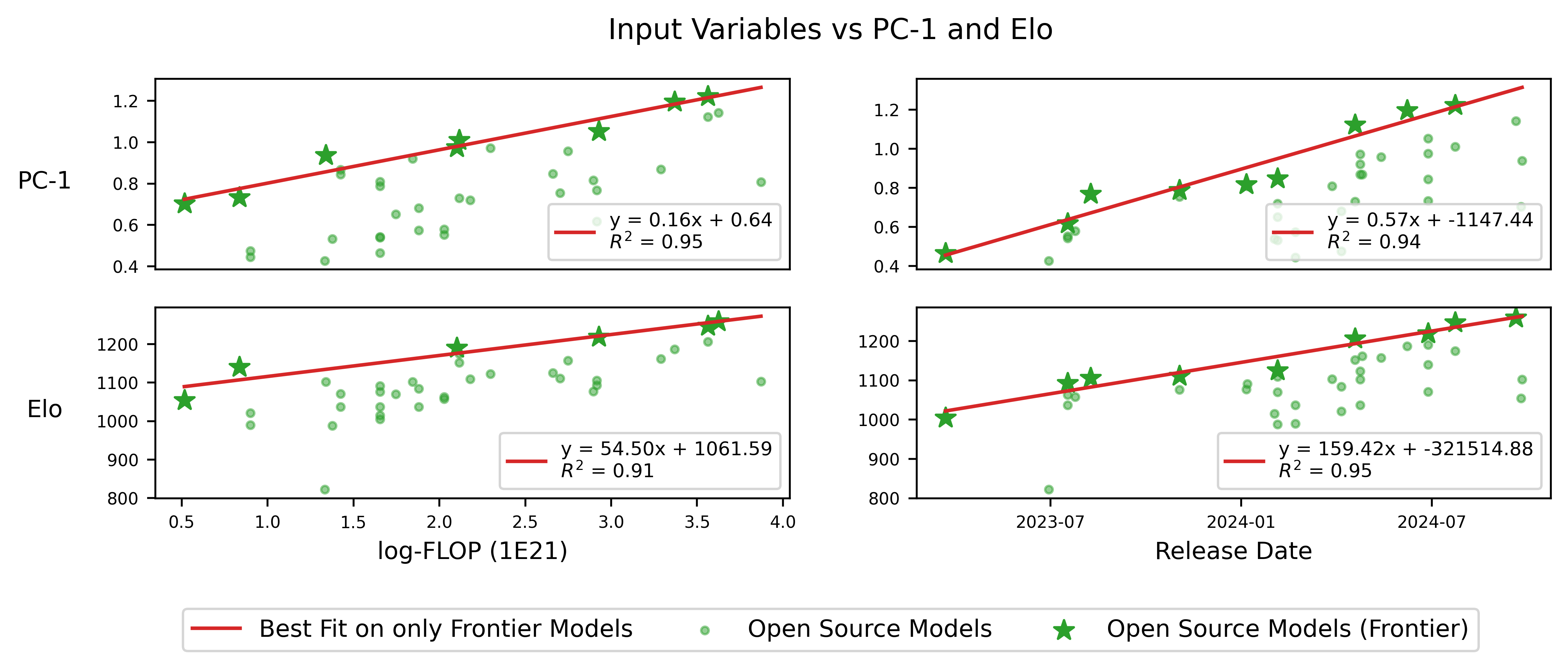}
    \vskip -0.1in
    \caption{
        Forecasting intermediate capability metrics from input variables for frontier models.  
        We find that both PC-1 and Elo are surprisingly linear when predicted from FLOP and release date, with all combinations having a $R^2$ $\ge$ 0.91.
    }
    \label{fig:methods}
\end{figure*}

Note that we only focus on pre-training FLOP and do not take into account post-training such as RLHF/RLAIF \citep{ouyang2022traininglanguagemodelsfollow, bai2022constitutionalaiharmlessnessai}, nor inference time compute. The ``inference compute'' paradigm started by OpenAI's o1 \citep{openai2024learningreasonllms} is not accounted for in our methodology. Column 1 of Figure \ref{fig:methods} depicts the relationship between scaled log-FLOP and intermediate capability metrics.

\textbf{Release date}: The date that the model was officially released for public use. While there is no inherent reason to assume that the release date influences the performance of the model, we speculate that it is a good aggregate measure of both algorithmic improvements \citep{ho2024algorithmic, xiao2024densinglawllms, erdil2023algorithmicprogresscomputervision} and increased training compute \citep{sevilla2022computetrends}, especially for frontier models. 

\subsubsection{Target Benchmarks}
\label{subsec:target_benchmark}

Target benchmarks are measures of downstream performance that compare the results of different models. In our specific case, we use benchmarks on the OpenLLM v2 leaderboard \citep{open-llm-leaderboard-v2} for backtesting. They include IFEval \citep{zhou2023instructionfollowingevaluationlargelanguage}, BBH \citep{suzgun2022challengingbigbenchtaskschainofthought}, MATH LVL 5 \citep{hendrycks2021measuringmathematicalproblemsolving}, GPQA \citep{rein2023gpqagraduatelevelgoogleproofqa}, MUSR \citep{sprague2024musrtestinglimitschainofthought}, and MMLU-PRO \citep{wang2024mmluprorobustchallengingmultitask}. We then use the method that we judge to be best in our comparisons to forecast future agent capabilities on three benchmarks: SWE-Bench Verified \cite{openai2024swebenchverified}, Cybench \cite{zhang2024cybenchframeworkevaluatingcybersecurity}, and RE-Bench \cite{wijk2024rebenchevaluatingfrontierai}.


\subsection{One-step Forecasting Approach}
``One-step'' approaches predict the benchmark score from an input variable directly.

First, we identify which models lie on the frontier for each choice of input variable and target benchmark. A point \((x_i, y_i)\) is on the frontier if there is no other point \((x_j, y_j)\) with \(x_j < x_i\) and \(y_j > y_i\). Visually, this corresponds to the upper boundary of points in Figure  \ref{fig:methods}.


Following \citet{ruan2024observational}, we assume a sigmoidal relationship between capability metrics and the eventual benchmark scores. 
Per input variable $x$ with frontier datapoints $x_i$, and corresponding benchmark scores $y_i$, we fit $a, b$.\footnote{We assume the ceiling is 1 for all benchmarks except RE-Bench. For RE-Bench, the sigmoid is scaled to asymptote at 1.67 instead of 1. This choice reflects the authors' estimate of the maximum achievable performance across their tasks  \cite{wijk2024rebenchevaluatingfrontierai}. Averaging those maximum performances yields 1.67, motivating our selection of this asymptote.}
\begin{equation}\label{eq:sigmoid}
    y_i = \sigma(a x_i + b)
\end{equation}


\subsection{Two-step Forecasting Approach}
For ``two-step'' approaches, we first predict an ``intermediate capability metric'' and then use that to predict the downstream performance.

There are a few reasons in favor of a two-step approach. First, it is conceptually plausible that there is some underlying ``general capability'' that is highly predictive of the output variables. 
Second, the intermediate capability metric might function as a ``regularization'' step, e.g. by projecting all training compute onto GPT-2-equivalent FLOP as shown in \citet{ruan2024observational} or onto Hoffmann scaling laws as shown in \citep{owen2024predictablelanguagemodelbenchmark}.
Third, we might have more data available for either of the two intermediate forecasts, thus improving the overall prediction.

\subsubsection{Intermediate Capability Metrics}
\label{subsec:capability_metrics}

We aim to reduce the complex behaviors of the model down to a single number as an underlying measure of the latent \emph{general capabilities}. We have three main criteria for selecting such a metric. First, we need the metric to be easily computable. Second, it should be available for a large range of models. Third, the metric has to have high predictive power for a large range of downstream benchmarks.

\textbf{PC-1}: \citet{ruan2024observational} show that around 80\% of the variance in benchmark scores is explained by the first principal component of a PCA on benchmarks. The PCA is applied to a model~$\times$~benchmark table. The authors show that as a result, PC-1 can be used as a measure of general capabilities and is predictive across a large variety of tasks. Note that \citet{ruan2024observational} use the top 3 components, whereas we limit ourselves to only the top one.

\textbf{Elo}: Chatbot Arena \citep{chatbot_arena} is an open platform for evaluating LLMs based on human preferences. Users are asked to compare responses from two models and select the one they prefer. The Elo rating system is then used to assign a score to each model based on the outcomes of these comparisons.

There are various advantages and disadvantages to each metric. Elo does not require extensive evaluation on benchmarks, which can be time-consuming, expensive, or use inconsistent methodologies. Elo ratings do not saturate, while PC-1 scores saturate when all of its component benchmarks are saturated.
However, Elo ratings are only meaningful in relationship to the exact rating pool in which they were calculated. As a result, any of our forecasts expressed in Elo is best interpreted as: “If a model had an Elo of X relative to today’s models, how good would it be?”—recognizing that, by the time the model is actually released, the entire rating landscape may have shifted.

\subsubsection{Fitting Two-step Approaches}


To fit the relationship between the input variable and the capability metric, we first determine which models are on the frontier as in the ``one-step'' approaches. Then, we fit a linear regression between the frontier points' input variable $x$ and the intermediate capability metric values $z$.
\begin{equation}\label{eq:linear}
    z_i = c x_i + d
\end{equation}
Figure~\ref{fig:methods} shows that a linear relationship between the input variable and the capability metric is usually a good fit.

In the second step, we fit a sigmoid between the capability metric and benchmark score.  
\begin{equation}\label{eq:sigmoid}
    y_j = \sigma(e z_j + f)
\end{equation}
We fit the sigmoid on \emph{all} available models instead of just the frontier since we expect that frontier models will have the same relationship between the underlying capability and benchmark score as other models. 

\section{Evaluating approaches through backtesting}

To compare all six of our approaches, we backtest them on existing data from the Open LLM Leaderboard v2 \citep{open-llm-leaderboard-v2} with six benchmarks: IFEval, BBH, MATH Lvl 5, GPQA, MUSR, and MMLU-PRO.
We only use the subset of Open LLM Leaderboard v2 that has Elo scores available, resulting in 38 models (see Appendix~\ref{app:models_on_both_leaderboards}).

However, before we determine which pathway is the most accurate overall, we want to compare the predictive power of input and intermediate variables.
In Section \ref{subsec:backtest_capability_metrics}, we backtest individual capability metrics to validate Elo as a potential candidate and compare it to PC-1, scaled log-FLOP, and release date.
Then, we move towards testing the entire pathway. In Section \ref{subsec:backtest_full_approaches}, we backtest all six full approaches.

\subsection{Backtesting capability metrics}
\label{subsec:backtest_capability_metrics}

\begin{figure*}[!htb]
    \centering
    \makebox[\columnwidth][c]{
    \includegraphics[width=1\textwidth]{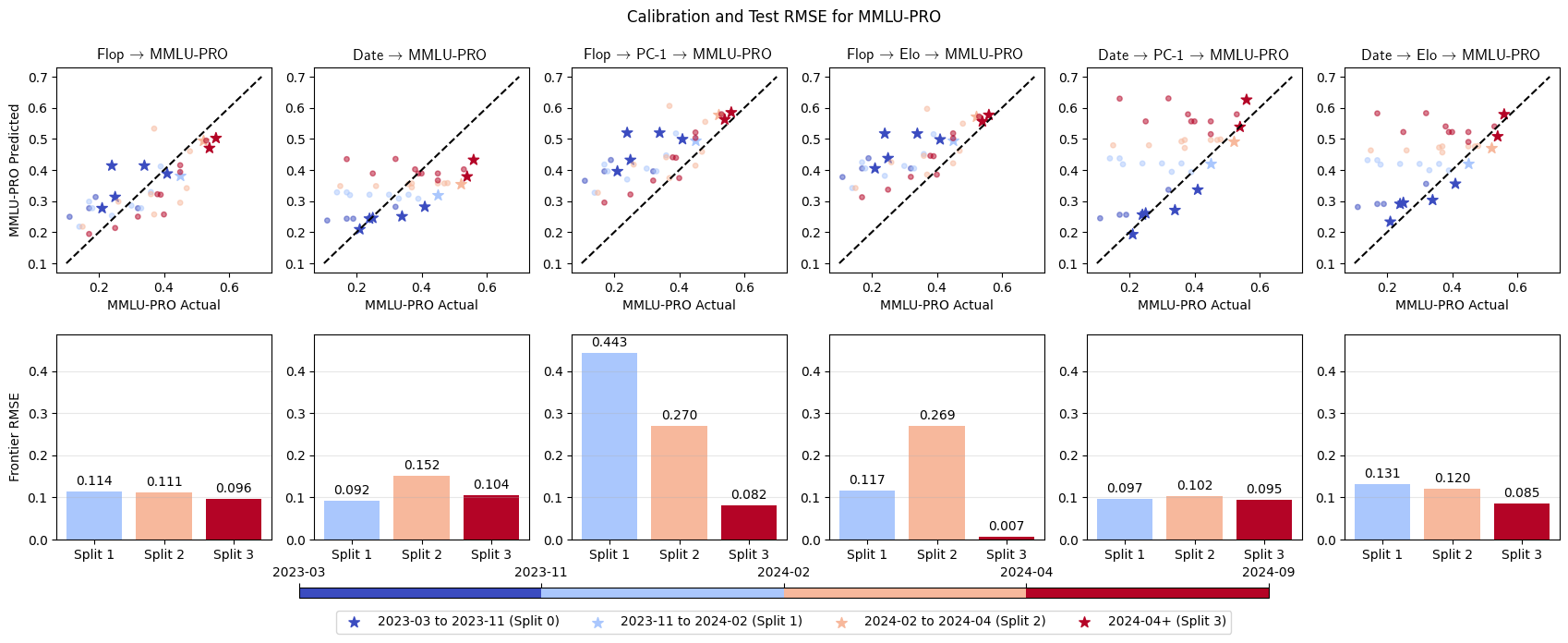}
    }
    \vskip -0.1in
    \caption{
        Visualization of backtesting forecasts for MMLU-PRO using the full method.
        We split the data into 4 parts with an equal number of models. We then fit a full path on split 1 and test on split 2, fit on 1 \& 2, and predict on 3, and so forth. 
        \textbf{Top:} Comparing predicted to actual performance. Frontier models are marked with stars. 
        \textbf{Bottom:} Average RMSE over frontier models. Bars are colored by the split they predict.
    }
    \label{fig:path_backtest}
\end{figure*}

To backtest capability metrics, we use expanding window cross-validation \cite{expanding_window_backtest} with 3 splits based on release date.
We first split our data up into 4 divisions with approximately equal model count.
Then, we train a statistical model only on the first split and evaluate it on the second split, another statistical model on the first and second split, and evaluate it on the third, and so on.
Our cross-validation methodology is displayed for predictions of MMLU-PRO with the full approach in Figure \ref{fig:path_backtest} (see also Section \ref{subsec:backtest_full_approaches}).

We are computing the error of only the capability metric, so we train just the sigmoid from the capability metric to the target benchmark. (Subplot 2 in Figure~\ref{fig:paths}).
For PC-1, we avoid testing on the training data and thus omit the benchmark we're predicting when fitting the PCA. Furthermore, we only use the data available up to that point when fitting the principal component vectors.  
To compare the overall performance of our four capability metrics, we compute the RMSE over our three splits and six benchmarks, for each approach.

\begin{table}[H]
    \begin{center}
        \begin{scriptsize}
            \setlength{\tabcolsep}{8.5pt}
            \begin{tabular}{lcccr}
                 \toprule
                 Capability Metric & PC-1 & Elo & log-FLOP & Release Date \\
                 \midrule
                 Test RMSE & 0.068 & 0.080 & 0.102 & 0.146 \\
                \bottomrule                     
            \end{tabular}
        \end{scriptsize}
    \end{center}
    \vskip -0.1in
    \caption{Average all-model test-split back-prediction RMSE for prediction of target benchmark from capability metrics. Intermediate metrics (PC-1 and Elo), outperform raw input variables (log-FLOP and Release Date).}
    \label{table:rmse_capability_metrics}
\end{table}

Table \ref{table:rmse_capability_metrics} displays the aggregated results. 
PC-1 performs best, followed by Elo, log-FLOP, and date as intermediate metrics. 
The full results of our capability metric backtesting can be found in Appendix~\ref{app:capability_metric_backtesting}.

\subsection{Backtesting full approaches}
\label{subsec:backtest_full_approaches}

To backtest the full paths, we use the same expanding window cross-validation procedure. 
However, there are two important differences. 
First, we are testing the complete path from input variable to benchmark score, ignoring the internal loss of the S-curve or linear regression subcomponents.
Second, we only compute error for data points on the frontier. If there are no frontier points in a split, that split is ignored.
We then aggregate the error in each split as usual. Since there are far fewer data points, the error is likely to be noisier.

\begin{table}[h]
   \begin{center}
       \begin{scriptsize}
           \setlength{\tabcolsep}{10pt}
           \begin{tabular}{l|cr}
                \toprule
                \textbf{Intermediate Variable} & \multicolumn{2}{c}{\textbf{Input Variable}}\\
                \cmidrule(r){1-1} \cmidrule(l){2-3}
                & log-FLOP & Date  \\
                \midrule
                \emph{None} (One-Step) &  0.119 & 0.125 \\
                Elo & 0.197 & 0.095 \\
                PC-1 & 0.105 & 0.082 \\
               \bottomrule
           \end{tabular}
       \end{scriptsize}
   \end{center}
   \vskip -0.1in
   \caption{Average frontier model test-split back-prediction RMSE for full approach. The path Date$\to$PC-1$\to$Benchmark performs best, followed by Date$\to$Elo$\to$Benchmark.}
   \label{table:rmse_paths}
\end{table}

Our results (see Table \ref{table:rmse_paths}) show that the best overall path is going from Release Date$\to$PC-1$\to$Benchmark, with an overall RMSE of 0.082, followed by Date$\to$Elo$\to$Benchmark with an RMSE of 0.095. Overall, using release date as the input variable outperforms log-FLOP.

\section{Predictions for Agentic Benchmarks}
\label{sec:agent_predictions}

Informed by our backtesting, we now want to apply the most suitable methodology to predict the performance of three LM agent benchmarks.  

\subsection{Choice of benchmarks}

First and foremost, we want the benchmarks to capture important, economically valuable skills such that our forecasts have meaningful real-world implications.
Second, we want to use benchmarks that have a high option space and require repeated interaction with the environment in order to measure agent capabilities rather than pure knowledge. 
Third, we want the benchmarks to be difficult but have easily verifiable solutions. 
Finally, we want them to be popular for general validation and to compare performance against other implementations.

As such, we use SWE-Bench Verified~\citep{jimenez2024swebenchlanguagemodelsresolve, openai2024swebenchverified}, where all problems have been human-verified and a public leaderboard exists, Cybench~\citep{zhang2024cybenchframeworkevaluatingcybersecurity}, which aims to be representative of real-world cybersecurity work, and RE-Bench~\citep{wijk2024rebenchevaluatingfrontierai}, which attempts to measure the AI R\&D capabilities of LM agents. METR has kindly shared scores for eight frontier models with us \cite{scorescore}.

\subsection{Forecasting methodology}
We only use release date as the input variable since training FLOP count is no longer publicly known for most frontier models.
Furthermore, we only use Elo as our capability metric since almost all publicly available frontier models are available on Chatbot Arena, but not necessarily all benchmark scores. 
In Section \ref{subsec:backtest_full_approaches}, we show that the release~Date$\to$Elo$\to$Benchmark~score path performs second-best in backtesting. Thus, it seems like a sufficiently good choice.

\subsection{Scaffolding}
We use the same scaffold for both SWE-Bench Verified and Cybench. For RE-Bench, we rely on METR's data, and thus don't have detailed knowledge of which scaffold was used.
Our scaffold attempts to be as simple as possible while avoiding simple known pitfalls.


We provide the model with three tools: a) A Bash shell, b) a Python shell, and c) a file editing tool that enables the model to view, create, and edit files by searching and replacing text and allows it to undo changes (similar to \citet{anthropic2024raisingbarswebench}).

All runs have a message cap of 50 messages and 2 million tokens. If the model runs out of context, we delete the earliest non-instruction messages.
Prompts for our scaffold are provided in Appendix \ref{app:scaffold}.

\subsection{Elicitation}
\label{subsec:elicitation}

The highest-performing scaffolds for each benchmark typically give more affordances to the model, or provide more inference-time compute. Furthermore, they often integrate prior knowledge about the benchmark into the scaffold, e.g. different prompts for isolating the bug, writing test cases, and retrying for SWE-Bench Verified.

Since our simple scaffold makes no use of additional inference compute, such as ``best-of-n'' or o1-style inference techniques \citep{openai2024learningreasonllms}, or highly task-specific prompts, we achieve a score of around 33\% on SWE-Bench Verified with Claude-Sonnet-3.5, while the best public scaffold known to be using Claude-Sonnet-3.5 on the SWE-Bench Verified leaderboard achieves 62.2\% \citep{pani2024sotaswebench}. 

Thus, we differentiate between a ``low-elicitation'' estimate, which should be seen as a general conservative estimate, and a ``high-elicitation'' estimate, which represents the best publicly known scaffolds at the time taken from publicly available leaderboards. The ``high-elicitation'' forecast has the advantage that it predicts the real public frontier, but the disadvantage that the scaffolds are almost always different between data points. 

We fit the low-elicitation forecast on only data gathered from our simple uniform scaffold. For the high-elicitation forecast, we combine \emph{all} data points, including both our own scaffold, and data from public leaderboards.  





\subsection{Results}

\begin{figure*}[!htb]
    \centering
    \includegraphics[width=\textwidth]{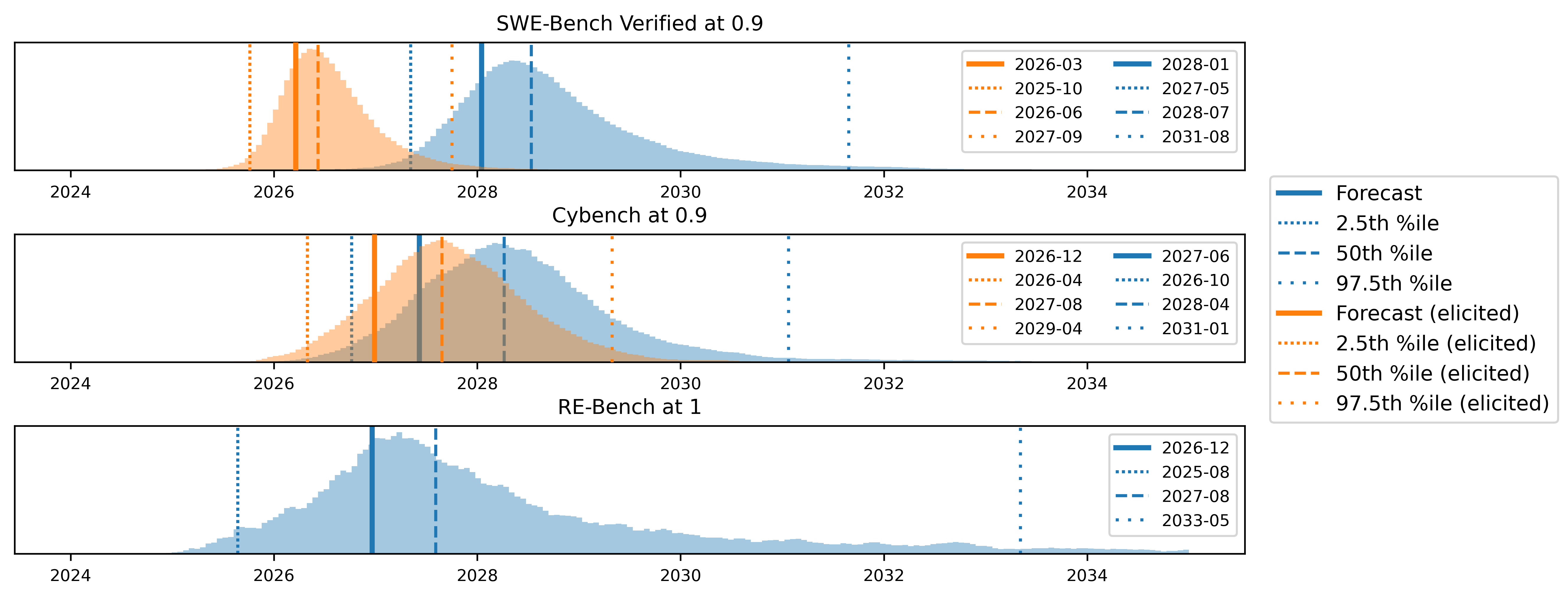}
    \vskip -0.1in
    \caption{
        Predictions for a 0.9 success rate on SWE-Bench Verified and Cybench and a score of 1 on RE-Bench for low and high elicitation, respectively. We compute the distribution using bootstrapping with 10,000 samples.
        Note that the medians (50th percentile) of these histograms do not necessarily equal the forecasts made with all data points in Figure \ref{fig:scaling-graph}.
    }
    \label{fig:distributions}
\end{figure*}

Figure \ref{fig:scaling-graph} shows the results of our forecasts until early 2027.

For SWE-Bench, we have access to all 17 models tested with our simple scaffold for the low elicitation effort and access to strong elicitation efforts of other groups from the public leaderboard.
Our model indicates that by January 2026, models with weak elicitation will achieve 54\% on SWE-Bench, and with better elicitation may achieve 87\%. However, our model does not take into account the potential for heavily increased test-time scaling, which may further increase performance.

Our forecast suggests that Cybench scores will be 55\% and 66\% in January 2026 for low and high-elicitation efforts, respectively. We observe that there is much less difference between the non-elicited and elicited cases, likely because far less effort has gone into eliciting Cybench performance to date.

On RE-Bench, we forecast a score of 0.73 by January 2026. Note that METR reported that they did not spend a lot of effort on elicitation, which suggests our estimates might be too conservative. Consequently, we exclude a high-elicitation scenario from our forecasts on this benchmark.


In Figure \ref{fig:distributions} we show the conditional distributions for a fixed benchmark score. We chose a score of 0.9 for SWE-Bench and Cybench as an arbitrary marker of strong performance and a score of 1 for RE-Bench, which is the expert baseline. 

With high elicitation, we expect SWE-Bench Verified to reach 90\% around March 2026, with a 95\% CI spanning from October 2025 to September 2027. With standard elicitation, we expect 90\% to be reached about two years later, in January 2028.

For Cybench, our best guess for high elicitation is December 2026, with a 95\% CI from April 2026 to April 2029. Standard elicitation predicts June 2027. 

Our forecast suggests that agent performance on RE-Bench may reach a score of 1—equivalent to the expert baseline reported by \citet{wijk2024rebenchevaluatingfrontierai}—around December 2026. We have much more uncertainty about this forecast, and our 95\% CI reflects this. It has a span of over 8 years, from August 2025 to May 2033. 

Across all three benchmarks and elicitation types, we observe that the probability distributions are asymmetric, with a longer right tail. This indicates greater uncertainty about potential delays compared to early achievements.

\section{Related Work}
\label{sec:related_work}

\paragraph{Pre-training scaling laws}
\citet{hestness2017deeplearningscalingpredictable} discuss power-law scaling of the error with respect to dataset size and parameter count across multiple domains.
\citet{kaplan2020scalinglawsneurallanguage} discusses specific power law exponents for training transformers optimally given a fixed compute budget, and demonstrate that these scaling laws hold over many orders of magnitude. These parameters are improved by \citet{hoffmann2022trainingcomputeoptimallargelanguage}, who introduce ``Chinchilla'' scaling laws. \citet{besiroglu2024chinchilla} repeat their experiment and further refine the parameters.

\paragraph{Predicting downstream performance}
\citet{Finnveden_2020} uses data from GPT-3 \citep{brown2020languagemodelsfewshotlearners} to extrapolate the performance of future models on downstream tasks, including Winograd, SuperGLUE, and ANLI.
\citet{owen2024predictablelanguagemodelbenchmark}, models BIG-Bench performance with respect to Scaled FLOP and finds that we can predict aggregated performance but not individual task performance.
\citet{ruan2024observational}, use a PCA on benchmark scores to decompose scores into a low-dimensional capability space. They find that performance across benchmarks can largely be explained by a single principle component which they interpret as a ``general capability factor''. \citet{polo2024sloth} extend this idea via a low-dimensional “skill” model that can predict across families if smaller models from the same family are available.  
\citet{zhang2024collaborativeperformancepredictionlarge}, applies collaborative filtering to handle sparse benchmark scores, i.e. cases where not all models have the same benchmark scores available. 

Less attention has been paid to directly predicting frontier performance instead of average performance. 
\citet{steinhardt2022ForecastingMLBenchmarks} discusses factors relevant to forecasting LLM performance, including training data availability and algorithmic progress. They attempt to predict frontier models' scores on MATH and MMLU. 
\citet{villalobos2024run} discusses how much training data future models are likely to have access to in further detail. 
\citet{ho2024algorithmic} discusses the rate of algorithmic progress, or how much improvement we are likely to see due to changes in architecture and improvements in the training process. 
\citet{cottier2024rising} discusses the increasing cost of training the underlying foundation models.
\citet{erdil2023power} discusses the base rate of breakthrough developments, and how it contributes to algorithmic progress.
\paragraph{Emergent capabilities}
Sudden jumps in capabilities on benchmark performance are often attributed to ``Emergent capabilities.'' 
\citet{srivastava2023imitationgamequantifyingextrapolating}, noticed that certain tasks in BIG-Bench experienced sudden jumps in capabilities.
\citet{wei2022emergentabilitieslargelanguage} discusses these emergent abilities in further detail.
\citet{schaeffer2023emergent} shows that emergent capabilities heavily depend on the choice of grading criterion or metric and that non-linear jumps in one metric can be smooth in another. 

\section{Discussion}
\label{sec:discussion}

\subsection{Limitations}
\label{sec:limitations}
\paragraph{Paradigm changes} 
While this paper does not make any explicit assumptions about the training paradigm of any particular model, we fit almost all predictions on models that were trained with the ``pre-training scaling'' paradigm, where the primary driver for downstream performance was improvements of pre-training. However, with OpenAI's o1 \citep{openai2024learningreasonllms}, we may start to see a new ``inference scaling'' paradigm where models are trained to utilize inference compute much more effectively through reasoning. This might invalidate our predictions and thus provide a reason to assume faster progress than our forecasts would suggest, even for high-elicitation predictions.
\paragraph{Underelicitation} 
As discussed in Section \ref{subsec:elicitation}, we did not put a lot of effort into elicitation.

As a consequence, we know that our results are significantly below frontier performance and that our ``low-elicitation'' predictions are conservative. Even the ``max-current-elicitation'' forecast might underestimate performance due to paradigm changes (see above) or later breakthroughs in agent scaffolding and elicitation.
\paragraph{Small sample size} 
Unfortunately, almost by definition, there are only a small number of frontier models. Therefore, our predictions have a small sample size. This is partially mitigated by making use of the two-step methodology and predicting the intermediate variable independently. However, we think the small sample size should imply large uncertainty about our forecasts. 

This limitation also affects our backtesting. Because our available test data is limited, we must rely on small evaluation windows, some as brief as two-month intervals. As a result, we have little empirical evidence regarding how our predictions might perform over longer periods.

\paragraph{Limited Scope of Evaluations}
The benchmarks we consider focus primarily on software engineering, cyber capabilities, and machine learning engineering. 
Noteworthy other agent benchmarks include GAIA \citep{mialon2023gaiabenchmarkgeneralai} and OS-World \citep{xie2024osworldbenchmarkingmultimodalagents} for browsing and tool use, as well as MLE-Bench \citep{chan2024mlebenchevaluatingmachinelearning} for additional machine learning capabilities.

%
%
\subsection{Future work}
The most straightforward way to extend this paper is by adding more agentic benchmarks and more models. 
Secondly, our ``high-elicitation'' estimates are done with very different scaffolds and techniques. We think there is a lot of room for improvement in choosing datapoints and designing forecasting techniques here. 
Thirdly, extending any forecasting technique with a factor for inference-time compute scaling might yield more accurate results. 

\section{Conclusion}
There are three primary novel contributions from this paper. First, we focus on predicting frontier performance instead of average performance. Second, we use different data types than previous work, e.g. using Elo as an intermediate variable and using release date as an input. One advantage of using multiple techniques is that we can choose methods based on the availability of data, e.g. for frontier models, release date is known, while training compute isn't. Third, we focus on benchmarks specifically designed for LM agents while previous work has often focused on QA benchmarks. 

\subsubsection*{Author Contributions}

Govind Pimpale and Axel H{\o}jmark ran all experiments and analyses. They also contributed substantially to the conceptual efforts. For example, AH came up with the idea of using Elo as an intermediate variable.
Jérémy Scheurer and Marius Hobbhahn co-supervised the project.
JS had the original idea for the project and developed the first roadmap.
GP and MH wrote the paper, supported by AH and JS.

\bibliography{bibliography}
\bibliographystyle{iclr2025_conference}

\newpage
\appendix
\onecolumn

\section{Scaled Compute Calculations}
\label{app:scaled_compute}

Recall that Hoffman loss is:
\begin{equation}
    \hat{L}(N, D) = E + \frac{A}{N^\alpha} + \frac{B}{D^\beta}
\end{equation}
The normalized scaled FLOP count $C_{opt}$ is:
\begin{align*}
    L_{\text{model}}        & = \hat{L}(N_{\text{model}}, D_{\text{model}})              \\
    N_{\text{opt}}, D_{\text{opt}} & = \text{hoffman\_optimal\_params}(L_{\text{model}}) \\
    C_{\text{opt}}          & = 6 N_{\text{opt}} D_{\text{opt}}                          \\
\end{align*}


Using the method of Lagrange multipliers, it can be shown that:
\begin{align*}
    N_{opt} &= \frac{A(\alpha + \beta)}{(l\beta)^\frac{1}{\alpha}} \\
    D_{opt} &= \frac{B(\alpha + \beta)}{(l\alpha)^\frac{1}{\beta}} \\ 
    \text{where} & \\
    l &= L_{budget} - E \\
\end{align*}

\section{Correlation between Elo and PC-1}

\begin{figure*}[h]
    \centering
    \includegraphics[width=\textwidth]{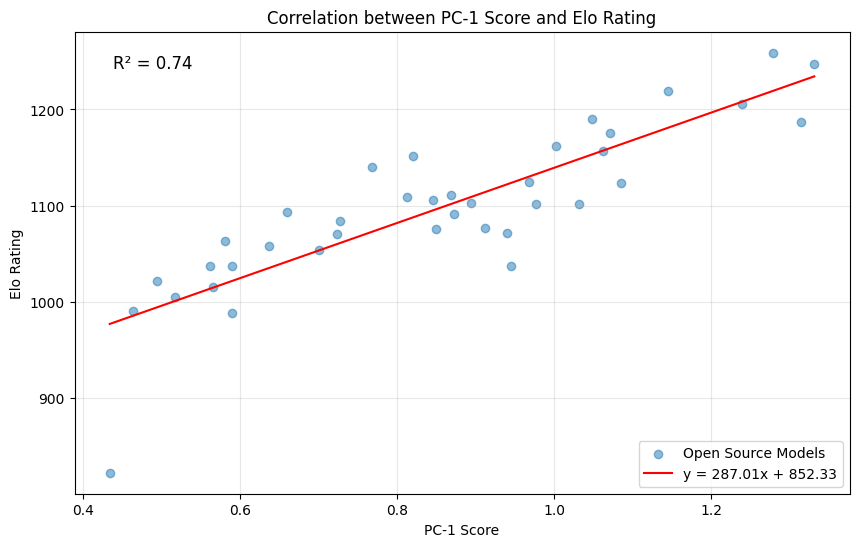}
    \vskip -0.1in
    \caption{Elo and PC-1 are well correlated, with an $R^2$ of 0.74} 
    \label{fig:elo-pc1-correlation}
\end{figure*}



\section{Capability Metric Backtesting Details}
\label{app:capability_metric_backtesting}

\begin{figure}[!htb]
     \centering
     \begin{subfigure}[b]{0.33\textwidth}
         \centering
        \includegraphics[width=\textwidth]{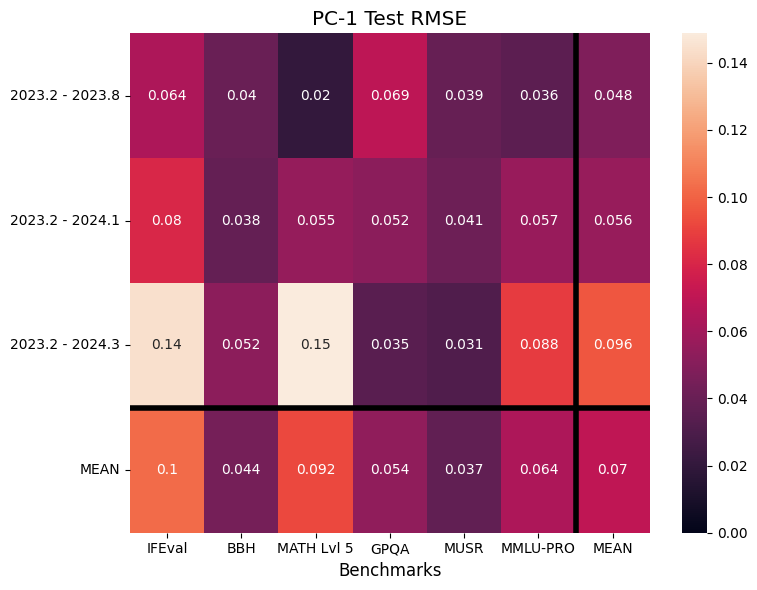}
        \caption{Test RMSE of PC-1}
        \label{fig:directpc1_perf_all}
     \end{subfigure}
     \hfill
     \begin{subfigure}[b]{0.33\textwidth}
        \centering
        \includegraphics[width=\textwidth]{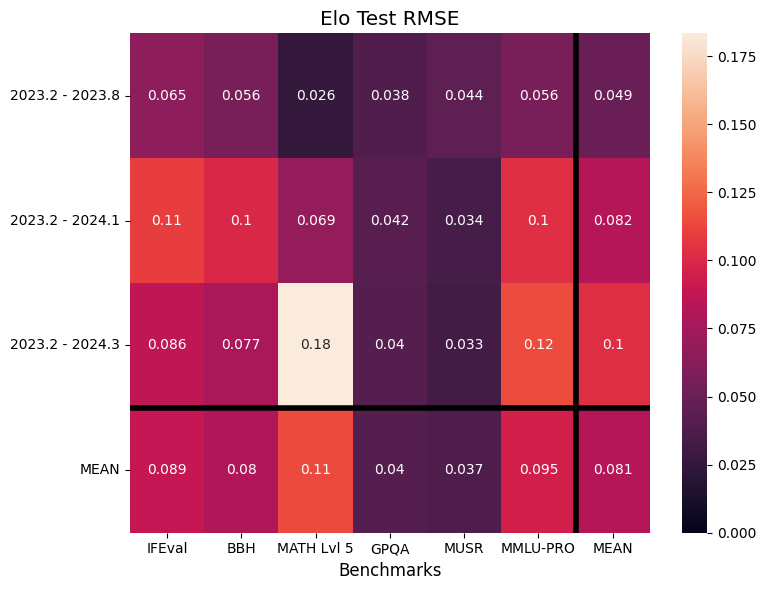}
        \caption{Test RMSE of Chatbot Arena Elo}
        \label{fig:directelo_perf_all}
     \end{subfigure}
     \hfill
     \begin{subfigure}[b]{0.33\textwidth}
         \centering
        \includegraphics[width=\textwidth]{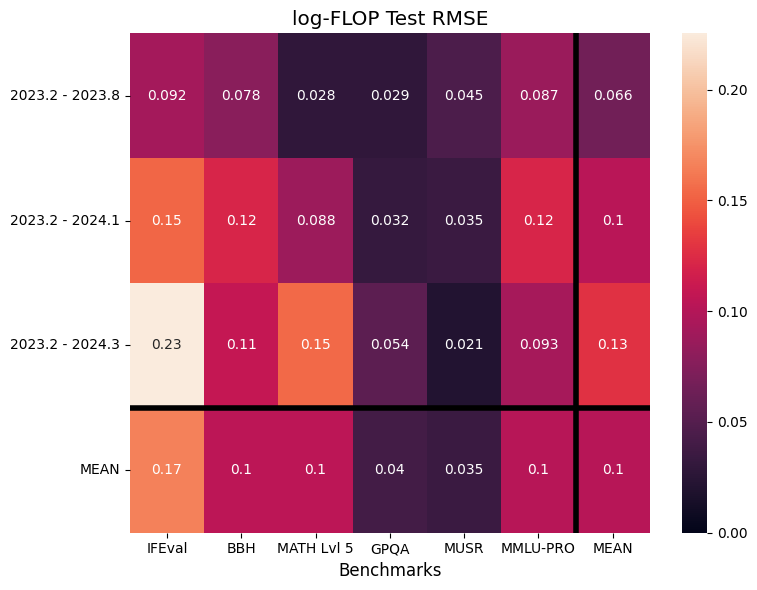}
        \caption{Test RMSE of log-FLOP    }
        \label{fig:directlogflop_perf_all}
     \end{subfigure}
    \caption{Full results of backtesting capability metrics}
    \label{fig:capability_metric}
\end{figure}

\section{Scaffold Details}
\label{app:scaffold}

\subsection{Prompts}

\begin{lstlisting}[basicstyle=\small,caption=Common System Prompt]
You are an autonomous AI agent. You are made up of a large language model, and a scaffolding program that runs on a linux VM. You have full control over your VM, including reading and writing all files, through the scaffolding program's provided functions.

You are state-of-the-art, and a capable strategist, writer, and programmer. You are resourceful and relentless, and you never give up. 
Never ask the user for help or guidance, as you are operating autonomously. There is no user at the other end of the conversation, only a linux VM.
Think step by step. Before you call a tool or command, reason out loud describing what command calls you are going to use and how they fit into your plan.
\end{lstlisting}

Note that we can only show this demonstration to agents who don't have native function calling
this is because we can't show examples of native function calls, as they are internal
\begin{lstlisting}[basicstyle=\small,caption=Demonstration of XML Function Calling]
Here is a demonstration of how to correctly accomplish another task.
It is included to show you how to correctly use the interface.
You do not need to follow exactly what is done in the demonstration.
**** DEMONSTRATION ****

**USER**:
YOUR TASK:
Find out how many days have passed since April 2nd 2019. Return just the number of days as an integer.

**ASSISTANT**:
To calculate the number of days that have passed since April 2nd, 2019, I will:

1. Get the current date.
2. Calculate the difference between the current date and April 2nd, 2019.
3. Return the result as an integer.

Let's start with getting the current date.

<bash>date +%F</bash>

**USER**:
BASH OUTPUT:
2024-07-25

**ASSISTANT**:
The current date is July 25th, 2024. Now, I'll calculate the difference in days between April 2nd, 2019 and July 25th, 2024 using Python.

Here is the step-by-step plan:
1. Define the start date (April 2nd, 2019).
2. Define the end date (July 25th, 2024).
3. Calculate the difference in days between the two dates.
4. Print the result.

Let's execute the Python code.

<bash>python3 -c "
from datetime import datetime

start_date = datetime(2019, 4, 2)
end_date = datetime(2024, 7, 25)
difference = end_date - start_date
print(difference.days)
"</bash>

**USER**:
BASH OUTPUT:
1941

**ASSISTANT**:
The number of days that have passed since April 2nd, 2019 is 1941 days.
<submit>1941</submit>

**USER**:
Successfully submitted answer.
**** END OF DEMONSTRATION ****
\end{lstlisting}

\subsection{Benchmark Specific Prompts}

We used a benchmark-specific prompt for SWE-Bench verified that aimed to give the model some more context on its task, without overfitting to specific properties of the problem

\begin{lstlisting}[basicstyle=\small,caption=Extra Prompt used for SWE-Bench Verified]
Below is a Github Issue that describes the problem you need to solve.
The repository has been cloned into /testbed (your current working directory). All dependencies are installed for you.
Your job is as follows:
1. Isolate the file/files where the bug is found.
2. Fix the bug.
3. Write a new test case that demonstrates the bug. The test case should be written in the same style as the existing tests.
4. Run the tests to ensure that the bug is fixed. Please ensure that you follow proper instructions to run tests, as described in the repo's documentation. Not all repositories have the same test running instructions. Note that some tests may fail, but as long as they are not related to the bug you fixed, you can ignore them.
5. Submit the empty string using the submit tool.

ALWAYS EXPLAIN YOUR REASONING BEFORE RUNNING A COMMAND. This will help you avoid mistakes and make it easier for us to understand your thought process.
\end{lstlisting}

For Cybench, we opted not to use a benchmark-specific prompt as the performance with the default instructions was adequate. 

\section{Agentic Data}
\label{app:agentic_data}
\begin{table}[H]
    \centering
    \begin{scriptsize}
        \begin{tabular}{lllllll}
            \toprule
            Model & Release Date & Chatbot Arena Elo & SWE-Bench Verified & Cybench & RE-Bench total \\
            \midrule
            openai/gpt-4o-2024-08-06 & 2024-08-06 & 1285 & 0.206 & 0.125 & 0.35501 \\
            openai/gpt-4o-mini-2024-07-18 & 2024-07-18 & 1274 & 0.078 & 0.05 & 0.19294 \\
            openai/gpt-4-turbo-2024-04-09 & 2023-11-06 & 1256 & 0.1706827309 & 0.05 & 0.27517 \\
            openai/gpt-3.5-turbo-0125 & 2022-03-15 & 1106 & 0.05 & 0.025 & \\
            openai/o1-mini-2024-09-12 & 2024-09-12 & 1306 & 0.182 & 0.15 & \\
            openai/o1-2024-12-17 & 2024-12-17 & 1335 & 0.342 & & 0.47043 \\
            anthropic/claude-3-5-sonnet-20241022 & 2024-10-22 & 1284 & 0.3306613226 & 0.2 & 0.50988 \\
            anthropic/claude-3-5-sonnet-20240620 & 2024-06-20 & 1268 & 0.2685370741 & 0.15 & 0.46139 \\
            anthropic/claude-3-sonnet-20240229 & 2024-02-29 & 1201 & 0.082 & 0.025 & 0.15334 \\
            anthropic/claude-3-opus-20240229 & 2024-02-29 & 1247 & 0.148 & 0.05 & 0.28037 \\
            anthropic/claude-2.1 & 2023-11-23 & & 0.038 & 0 & \\
            together/meta-llama--Meta-Llama-3.1-8B-Instruct-Turbo & 2024-07-23 & 1176 & 0.0120240481 & 0.025 & \\
            together/meta-llama--Meta-Llama-3.1-70B-Instruct-Turbo & 2024-07-23 & 1248 & 0.116 & 0.05 & \\
            together/meta-llama--Meta-Llama-3.1-405B-Instruct-Turbo & 2024-07-23 & 1268 & 0.164 & 0.075 & \\
            together/Qwen--Qwen2.5-7B-Instruct-Turbo & 2024-09-19 & & 0 & 0 & \\
            together/Qwen--Qwen2.5-Coder-32B-Instruct & 2024-09-19 & & 0 & 0.05 & \\
            together/Qwen--Qwen2.5-72B-Instruct-Turbo & 2024-09-19 & 1258 & 0.124 & 0.1 & \\
            \bottomrule
        \end{tabular}
    \end{scriptsize}
    \caption{Data collected for Cybench, SWE-Bench Verified, and RE-Bench}
    \label{table:agentic_data}
\end{table}

\section{Leaderboard Data}
\label{app:models_on_both_leaderboards}
\begin{table}[H]
    \centering
\begin{tiny}
    \begin{tabular}{llrrrrrrrrrr}
    \toprule
     & model & Elo & IFEval & BBH & MATH Lvl 5 & GPQA & MUSR & MMLU-PRO & release\_date & N (10e9) & D (10e12) \\
    \midrule
    0 & Qwen2.5-72B-Instruct & 1259 & 0.86 & 0.73 & 0.01 & 0.38 & 0.42 & 0.56 & 2024.72 & 72.00 & 18.00 \\
    1 & Meta-Llama-3.1-70B-Instruct & 1247 & 0.87 & 0.69 & 0.31 & 0.36 & 0.46 & 0.53 & 2024.56 & 70.00 & 15.00 \\
    2 & Gemma-2-27B-it & 1219 & 0.80 & 0.65 & 0.01 & 0.38 & 0.40 & 0.45 & 2024.49 & 27.00 & 13.00 \\
    3 & Command R+ (08-2024) & 1215 & 0.75 & 0.60 & 0.12 & 0.35 & 0.48 & 0.44 & 2024.65 & 104.00 & NaN \\
    4 & Llama-3-70B-Instruct & 1206 & 0.81 & 0.65 & 0.25 & 0.29 & 0.42 & 0.52 & 2024.30 & 70.00 & 15.00 \\
    5 & Gemma-2-9B-it & 1190 & 0.74 & 0.60 & 0.00 & 0.36 & 0.41 & 0.39 & 2024.49 & 9.00 & 8.00 \\
    6 & Qwen2-72B-Instruct & 1187 & 0.80 & 0.70 & 0.38 & 0.37 & 0.46 & 0.54 & 2024.44 & 72.00 & 7.00 \\
    7 & Meta-Llama-3.1-8B-Instruct & 1175 & 0.79 & 0.51 & 0.19 & 0.27 & 0.39 & 0.38 & 2024.56 & 8.00 & 15.00 \\
    8 & Qwen1.5-110B-Chat & 1162 & 0.59 & 0.62 & 0.00 & 0.34 & 0.45 & 0.48 & 2024.32 & 110.00 & 3.00 \\
    9 & 01-ai/Yi-1.5-34B-Chat & 1157 & 0.61 & 0.61 & 0.25 & 0.36 & 0.43 & 0.45 & 2024.37 & 34.00 & 3.60 \\
    10 & Llama-3-8B-Instruct & 1152 & 0.48 & 0.49 & 0.09 & 0.29 & 0.38 & 0.36 & 2024.30 & 8.00 & 15.00 \\
    11 & internlm/internlm2\_5-20b-chat & 1149 & 0.70 & 0.75 & 0.00 & 0.32 & 0.46 & 0.40 & 2024.30 & 20.00 & NaN \\
    12 & Mixtral-8x22b-Instruct-v0.1 & 1148 & 0.72 & 0.61 & 0.19 & 0.37 & 0.43 & 0.45 & 2024.30 & 141.00 & NaN \\
    13 & Gemma-2-2b-it & 1140 & 0.57 & 0.42 & 0.00 & 0.27 & 0.39 & 0.25 & 2024.49 & 2.00 & 2.00 \\
    14 & HuggingFaceH4/zephyr-orpo-141b-A35b-v0.1 & 1127 & 0.65 & 0.63 & 0.20 & 0.38 & 0.45 & 0.46 & 2024.28 & 141.00 & NaN \\
    15 & Qwen1.5-32B-Chat & 1125 & 0.55 & 0.61 & 0.07 & 0.31 & 0.42 & 0.45 & 2024.10 & 32.00 & 3.00 \\
    16 & microsoft/Phi-3-medium-4k-instruct & 1123 & 0.64 & 0.64 & 0.18 & 0.34 & 0.43 & 0.47 & 2024.31 & 14.00 & 4.90 \\
    17 & Mixtral-8x7B-Instruct-v0.1 & 1114 & 0.56 & 0.50 & 0.09 & 0.30 & 0.42 & 0.37 & 2023.95 & 47.00 & NaN \\
    18 & 01-ai/Yi-34B-Chat & 1111 & 0.47 & 0.56 & 0.05 & 0.34 & 0.40 & 0.41 & 2023.84 & 34.00 & 3.10 \\
    19 & Qwen1.5-14B-Chat & 1109 & 0.48 & 0.52 & 0.00 & 0.27 & 0.44 & 0.36 & 2024.10 & 14.00 & 3.00 \\
    20 & WizardLM/WizardLM-70B-V1.0 & 1106 & 0.50 & 0.56 & 0.04 & 0.27 & 0.44 & 0.34 & 2023.61 & 70.00 & 2.00 \\
    21 & DBRX-Instruct-Preview & 1103 & 0.54 & 0.54 & 0.07 & 0.34 & 0.43 & 0.37 & 2024.24 & 132.00 & 12.00 \\
    22 & Meta-Llama-3.2-3B-Instruct & 1102 & 0.74 & 0.46 & 0.17 & 0.28 & 0.35 & 0.32 & 2024.74 & 3.00 & 9.00 \\
    23 & microsoft/Phi-3-small-8k-instruct & 1102 & 0.65 & 0.62 & 0.03 & 0.31 & 0.46 & 0.45 & 2024.31 & 7.00 & 4.90 \\
    24 & meta-llama/Llama-2-70b-chat-hf & 1093 & 0.50 & 0.30 & 0.01 & 0.26 & 0.37 & 0.24 & 2023.55 & 70.00 & 2.00 \\
    25 & openchat/openchat-3.5-0106 & 1091 & 0.60 & 0.46 & 0.07 & 0.31 & 0.43 & 0.33 & 2024.02 & 7.00 & 2.00 \\
    26 & berkeley-nest/Starling-LM-7B-alpha & 1088 & 0.55 & 0.44 & 0.08 & 0.30 & 0.41 & 0.32 & 2023.87 & 7.00 & NaN \\
    27 & google/gemma-1.1-7b-it & 1084 & 0.50 & 0.39 & 0.04 & 0.29 & 0.42 & 0.26 & 2024.26 & 7.00 & 6.00 \\
    28 & NousResearch/Nous-Hermes-2-Mixtral-8x7B-DPO & 1084 & 0.59 & 0.55 & 0.12 & 0.32 & 0.46 & 0.37 & 2024.04 & 47.00 & NaN \\
    29 & deepseek-ai/deepseek-llm-67b-chat & 1077 & 0.56 & 0.52 & 0.07 & 0.32 & 0.51 & 0.39 & 2024.01 & 67.00 & 2.00 \\
    30 & openchat/openchat\_3.5 & 1076 & 0.59 & 0.44 & 0.07 & 0.30 & 0.42 & 0.32 & 2023.84 & 7.00 & 2.00 \\
    31 & teknium/OpenHermes-2.5-Mistral-7B & 1074 & 0.56 & 0.49 & 0.05 & 0.28 & 0.42 & 0.31 & 2023.83 & 7.00 & NaN \\
    32 & mistralai/Mistral-7B-Instruct-v0.2 & 1072 & 0.55 & 0.45 & 0.03 & 0.28 & 0.40 & 0.27 & 2023.95 & 7.00 & NaN \\
    33 & microsoft/Phi-3-mini-4k-instruct & 1071 & 0.55 & 0.55 & 0.15 & 0.33 & 0.43 & 0.40 & 2024.49 & 3.80 & 4.90 \\
    34 & Qwen1.5-7B-Chat & 1070 & 0.44 & 0.45 & 0.00 & 0.30 & 0.38 & 0.30 & 2024.10 & 7.00 & 3.00 \\
    35 & meta-llama/Llama-2-13b-chat-hf & 1063 & 0.40 & 0.33 & 0.01 & 0.23 & 0.40 & 0.19 & 2023.55 & 13.00 & 2.00 \\
    36 & upstage/SOLAR-10.7B-Instruct-v1.0 & 1062 & 0.47 & 0.52 & 0.00 & 0.31 & 0.39 & 0.31 & 2023.95 & 10.70 & NaN \\
    37 & WizardLM/WizardLM-13B-V1.2 & 1058 & 0.34 & 0.45 & 0.02 & 0.26 & 0.44 & 0.25 & 2023.56 & 13.00 & 2.00 \\
    38 & Meta-Llama-3.2-1B-Instruct & 1054 & 0.57 & 0.35 & 0.03 & 0.28 & 0.33 & 0.17 & 2024.73 & 1.00 & 9.00 \\
    39 & HuggingFaceH4/zephyr-7b-beta & 1053 & 0.50 & 0.43 & 0.03 & 0.29 & 0.39 & 0.28 & 2023.82 & 7.00 & NaN \\
    40 & HuggingFaceH4/zephyr-7b-alpha & 1041 & 0.52 & 0.46 & 0.02 & 0.30 & 0.39 & 0.28 & 2023.79 & 7.00 & NaN \\
    41 & google/gemma-7b-it & 1037 & 0.39 & 0.36 & 0.02 & 0.28 & 0.43 & 0.17 & 2024.14 & 7.00 & 6.00 \\
    42 & Phi-3-Mini-128k-Instruct & 1037 & 0.60 & 0.56 & 0.10 & 0.32 & 0.39 & 0.37 & 2024.31 & 3.80 & 4.90 \\
    43 & meta-llama/Llama-2-7b-chat-hf & 1037 & 0.40 & 0.31 & 0.01 & 0.25 & 0.37 & 0.17 & 2023.55 & 7.00 & 2.00 \\
    44 & google/gemma-1.1-2b-it & 1021 & 0.31 & 0.32 & 0.00 & 0.27 & 0.34 & 0.15 & 2024.26 & 2.00 & 3.00 \\
    45 & allenai/OLMo-7B-Instruct & 1015 & 0.35 & 0.37 & 0.01 & 0.27 & 0.38 & 0.18 & 2024.09 & 7.00 & 2.00 \\
    46 & mistralai/Mistral-7B-Instruct-v0.1 & 1008 & 0.45 & 0.34 & 0.02 & 0.25 & 0.38 & 0.24 & 2023.74 & 7.00 & NaN \\
    47 & lmsys/vicuna-7b-v1.5 & 1005 & 0.24 & 0.39 & 0.01 & 0.26 & 0.42 & 0.21 & 2023.22 & 7.00 & 2.00 \\
    48 & google/gemma-2b-it & 990 & 0.27 & 0.32 & 0.00 & 0.28 & 0.33 & 0.14 & 2024.14 & 2.00 & 3.00 \\
    49 & Qwen1.5-4B-Chat & 988 & 0.32 & 0.40 & 0.01 & 0.27 & 0.40 & 0.24 & 2024.10 & 4.00 & 3.00 \\
    50 & databricks/dolly-v2-12b & 822 & 0.24 & 0.33 & 0.01 & 0.24 & 0.37 & 0.11 & 2023.50 & 12.00 & 0.30 \\
    \bottomrule
    \end{tabular}
    \end{tiny}
    
    \caption{Data collected from OpenLLM Leaderboard 2 and Chatbot Arena. Note that we only use the subset with no NaN values, ultimately resulting in 38 models.}
    \label{table:leaderboard_data}
\end{table}

\end{document}